\documentclass[10pt,twocolumn,letterpaper]{article}

\usepackage{cvpr}
\usepackage{times}
\usepackage{epsfig}
\usepackage{graphicx}
\usepackage{amsmath}
\usepackage{amssymb}
\usepackage{lipsum}
\usepackage[linesnumbered,ruled,boxed ]{algorithm2e}
\SetKwComment{Comment}{$\triangleright$\ }{}

\usepackage[breaklinks=true,bookmarks=false]{hyperref}

\cvprfinalcopy 

\def\cvprPaperID{****} 

\begin{document}

\title{Detecting Kissing Scenes in a Database of Hollywood Films}

\author{Amir Ziai\\
Stanford University \\
Stanford, CA \\
{\tt\small amirziai@stanford.edu}
}

\maketitle



\section{Abstract}
Detecting scene types in a movie can be very useful for application such as video editing, ratings assignment, and personalization. We propose a system for detecting kissing scenes in a movie. This system consists of two components. The first component is a binary classifier that predicts a binary label (i.e. kissing or not) given a features exctracted from both the still frames and audio waves of a one-second segment. The second component aggregates the binary labels for contiguous non-overlapping segments into a set of kissing scenes. We experimented with a variety of 2D and 3D convolutional architectures such as ResNet, DesnseNet, and VGGish and developed a highly accurate kissing detector that achieves a validation F1 score of 0.95 on a diverse database of Hollywood films ranging many genres and spanning multiple decades. The code for this project is available at \href{http://github.com/amirziai/kissing-detector}{http://github.com/amirziai/kissing-detector}.
\section{Introduction}

Detecting scenes in videos has not achieved the same level of success as other important computer vision tasks such as object classification. Accurate scene detectors can be used to enrich the video's metadata with scene types and segments that can be easily searched and retrieved by users. Such annotations are pivotal in tasks such as video editing, determining movie ratings, and personalization. Most existing systems either classify still frames or recognize actions in the entire video, which may or may not be present at every scene. In this work we will propose a system for detecting and extracting kissing scenes in a movie.

The motivation for this problem comes from the research conducted by Poms et al \cite{poms2018scanner}. In their work the authors have developed a system for large-scale video synthesis using compositions of spatiotemporal labels. These labels are extracted from a number of annotators such as face bounding boxes, face embeddings, gender classifier, and so on. Leveraging these labels the authors have constructed a query language for extracting segments that adhere to a heuristic. For example one can express kissing as a query that involves faces being very close to each other. However this methodology has turned out to be unsuccessful for detecting kissing as faces can be occluded and closeness of faces does not necessarily correlate with kissing.

The input to our system is a movie and the output is non-overlapping kissing segments within the movie. For example for a 60 minute movie $M$ with two one-minute kissing scenes at minutes 5 and 55 the system is expected to return $K_1$ and $K_2$ where $K_1$ is the first kissing segment and $K_2$ is the second.

The first step is to chop up the video into non-overlapping segments of one second. We extract two sets of features from each one-second segment. The first feature set is extracted from the last still and is passed to a ResNet-18. We have detached the last fully connected layer and capture the 512 dimensional output. The second set is extracted from the audio and is fed to a VGG-like architecture called VGGish which produces a 128 dimensional embedding. We then concatenate the 512 and 128 dimensional outputs into a 640 dimensional tensor which is fed to a fully connected linear layer, which then outputs a binary class of either kissing or not kissing.

These one-second predictions are then passed to a segmentor algorithm for finding long non-overlapping segments, where the majority of one-second segments are predicted to be kissing.

\section{Related work}

End-to-end approaches for video classification using raw pixels values have been shown to be highly successful in the past few years.

Karpathy et al. \cite{karpathya2014large} showed that specialized architectures can be employed to exploit temporal information alongside spatial information which are typically derived using convolutional networks. The authors used such architectures to classify 1M YouTube videos into one of 487 classes. This strategy is effective for categorizing short videos but detecting certain types of scenes within a film requires a more granular approach. Yu et al. \cite{yu2019violent} developed a violent scene detection algorithm by combining three-dimensional histograms of gradient orientation (HOG3D), bag of visual words, and feature pooling. These features are passed to a kernel extreme learning machine (KELM) and a binary label is produced.

Another approach is to use two stream networks. Simonyan et al \cite{simonyan2014two} explicitly modeled the temporal featurs in a stack of optical flow vectors in parallel to a network for capturing the spatial context. These networks are trained separately and then combined using an SVM. The predicted output is averaged across the sampled frames. This approach suffers from the false label assignment problem. In other words, the label is applied to all the sampled frames, and the labeled action may not be represented in all frames.

More recently, 3D convolutional networks have shown ImageNet-level success for video classification. Hara et al. \cite{hara2018can} have shown that these models can retrace the effectiveness of 2D networks and can be as deep (e.g. 152 layers for ResNet). 3D convnets can be resource intensive for training and inference, however, they present an elegant framework for processing video data. Carreira et al. \cite{carreira2017quo} extend the two stream work with 3D networks and leverage pre-trained 2D models by repeating the weights in the third dimension. Our approach is closest to this line of work and explores using parallel networks (2D and 3D) consuming image and audio features which is, to the best of our knowledge, a novel approach.

\section{Methods}
\subsection{Input and output}
The input is a single video and the output is a set of non-overlapping kissing scenes detected within the video. Our proposed system achieves this using two components.

The first component is a binary classification model that takes contiguous non-overlapping one-second segment of the video and predicts a binary label for each (i.e. kissing or not). The second component takes the predictions from these segments and aggregates them to a set of kissing scenes. Figure 1 depicts this high-level process. We will explain each of these components in turn.

\begin{figure}[t]
\begin{center}
  \includegraphics[width=0.8\linewidth]{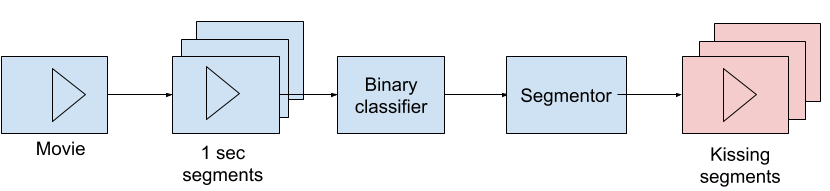}
\end{center}
   \caption{Kissing detector}
\label{fig:long}
\label{fig:onecol}
\end{figure}

\subsection{Binary classifier}
The binary classification component consists of two architectures. The first architecture is an 18-layer ResNet CNN that operates on the last frame in the one-second segment in the form of a 3-channel 224x224 tensor. We have detached the last fully connected layer and use the 512 dimensional output from the preceding layer. We use the pre-trained weights extracted from a ResNet-18 model trained on ImageNet which is available through PyTorch Hub\cite{paszke2017automatic}.

The second architecture is a VGG-like \cite{simonyan2014very} architecture called VGGish that operates on a transformation of the audio wave from the last 960ms of the one-second segment. This transformation is a single channel 96x64 tensor. VGGish is a convolutional network which effectively treats the transformed audio as if it were an image and generates a semantically meaningful 128 dimensional embedding. We used the pre-trained weights \cite{hershey2017cnn} for VGGish obtained from training the architecture on the AudioSet dataset \cite{gemmeke2017audio}.

The details of the still and audio feature extraction and transformation logic is provided in the upcoming Dataset and Features section.

The 512 and 128 dimensional outputs from ResNet and VGGish are concatenated into a 640 dimensional tensor and fed into a linear layer for generating binary labels. Figure 2 illustrates this component.

\begin{figure}[t]
\begin{center}
  \includegraphics[width=0.8\linewidth]{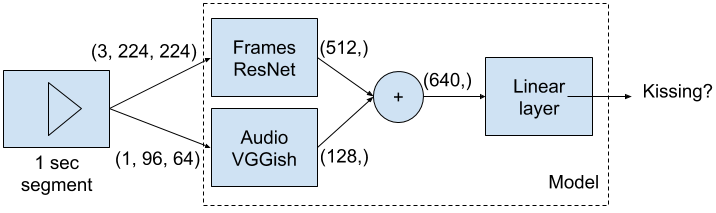}
\end{center}
   \caption{Binary classifier component}
\label{fig:long}
\label{fig:onecol}
\end{figure}

We train this model using cross entropy loss aggregated over all training examples. The loss for each training example is computed as follows:
$$\mathcal{L} (\hat{y}, c) = -\log (\dfrac{ \exp(\hat{y}[c]) }{ \exp(\hat{y}[0]) + \exp(\hat{y}[1]) }) $$
where $c \in \{0, 1 \}$ is the class label and $\hat{y}[c]$ is the unnormalized score for class $c$ generated by the final linear layer.

Throughout this work we've used an Adam optimizer learning rate of 0.001, $\beta_1=0.9$ and $\beta_2=0.999$ \cite{kingma2014adam}.

\subsubsection{ResNet}
We use ResNet-18 as described in \cite{he2016deep}. By introducing identity shortcut connections the network can effectively skip one or more layer which makes training deeper networks easier without an increase in the network complexity or the number of parameters.

The model architecture for ResNet follows a similar pattern to that of a VGG network \cite{simonyan2014very} of the same depth. Most of the convolutional layers have 3x3 filters and downsampling is performed by using convolutional layers of stride 2.

\subsubsection{VGGish}
The VGGish architecture \cite{hershey2017cnn} has been shown to be a very effective feature extractor for downstream Acoustic Event Detection (AED) detection task \cite{gemmeke2017audio} and we will use it for incorporating information extracted from audio of each one-second segment.

The network consists of a set of convolutional layers followed by max pooling that downsample the input starting from 64 channels and ending at 512. The 512-channel 6x4 tensor is then flattened and passed through a linear layer that outputs a 4096 dimensional output. Finally another fully connected layer of the same size as applied and then the last linear layer outputs the 128 dimensional output which we use as our audio features.

\subsubsection{3D ResNet}
We have also explored a three-dimensional version of ResNet-34 \cite{hara2018can}. This variant uses 3x3x3 kernels. The spatial input is still a 3-channel 224x224 image and 16 temporal frames are included. This allows the architecture to capture motion by incorporating temporal information \cite{hara2017learning}. However, this architecture has more features and a higher memory footprint per batch. In our experiments the training time per batch was increased by a factor of 10, some of which can be explained by using a deeper network relative to ResNet-18 which we used with 2D convolutions.

\subsection{Segmentor}
The second component of our system is an aggregation algorithm that combines the list of predicted labels $P$ from the binary classifier and generates a list of kissing segmented. For instance, consider a 60 minute movie with a two minute kissing scene starting at minute 30. The classifier will output 3600 predictions which we'll capture in $P$. The expected output from the segmentor, assuming a perfect classifier, is a list with a single segment starting at minute 30 and ending at minute 32.

The aggregation logic is detailed in Algorithm 1. The pruning step loops compares each segments to all the other segments and removes the smaller of 

\begin{algorithm}
    \SetKwInOut{Input}{Input}
    \SetKwInOut{Output}{Output}

    \underline{function segmentor} $(P, m, t)$\;
    \Input{$P$ is a list of 1-second segment predictions \newline
    $m$ is the minimum length of a segment in seconds (e.g. 10 seconds) \newline
    $t$ is the minimum fraction of kissing frames within the segment for it to qualify (e.g. 0.7)}
    \Output{List of kissing segments}
    C = [] \Comment*[r]{Empty list of segments}
    \For{$i = 1$ to $\lvert P \lvert$}{
        \If{$P[i] == 1$}{
            \For{$j = i + m + 1$ to $\lvert P \lvert$}{ 
                \If{$P[j] == 1$}{
                    $S := $ segment defined by $i$ and $j$ \;
                    $K := $ frames in $S$ where $s$ is the frame index and $P[s] == 1$ \;
                    $S_t = \lvert K \lvert / \lvert S \lvert$ \;
                    \If{$S_t \ge t$ }{
                    C.append($S$)
                    }
                }
            }
        }
      }
    \While{overlapping segments exist in $C$}{
        \For{($s_1$, $s_2$) in $C \times C$}{
            \If{$s_1 \neq s_2$ and $s_1$ overlaps with $s_2$}{
                Remove smaller of $s_1$ and $s_2$
            }
        }
    }
    return $C$
    \caption{Find non-overlapping kissing segments}
\end{algorithm}

\section{Dataset and Features}
The data we are using is a 2.3TB database of 600 Hollywood films spanning 1915 to 2016 and covering a broad range of genres and resolutions. These files can range between 200MB and 12GB.

\subsection{Pre-processing pipeline}
We have hand-selected a subset of 100 movies and then annotated kissing segments in these movies. The non-annotated segments are considered non-kissing segments and labeled as such in the dataset as shown in Figure 4. We have produced a total of 263 kissing segments and 363 non-kissing segments ranging 10 to 120 seconds in length. The dataset is split into train, validation, and test partitions with 80\%, 10\% and 10\% proportions respectively.

\begin{figure}[t]
\begin{center}
  \includegraphics[width=0.8\linewidth]{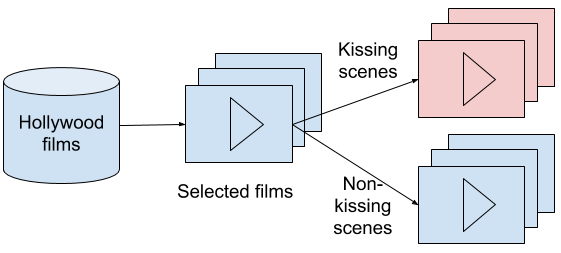}
\end{center}
   \caption{Data annotation}
\label{fig:long}
\label{fig:onecol}
\end{figure}

\subsection{Features}
For each movie annotated segment we extract two sets of features. Figure 3 depicts the parallel still and audio extraction for a hypothetical N-second video.

\begin{figure}[t]
\begin{center}
  \includegraphics[width=0.8\linewidth]{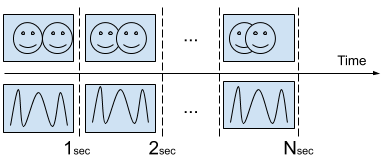}
\end{center}
   \caption{Parallel still and audio extraction per 1-second segment}
\label{fig:long}
\label{fig:onecol}
\end{figure}

\subsubsection{Image features}
The image features are extracted at the end of each one-second segment. In the training pipeline we randomly crop a 224x224 image from the still frame, randomly sampled from patches whose size is uniformly distributed between 8\% and 100\% of the image area and with aspect ratio uniformly selected between 3/4 and 4/3 (same as in \cite{szegedy2015going}). We then perform a random horizontal flip and then standardize each channel using the mean and standard deviation computed over the entire dataset. At test time we apply the same transformations except the for the random horizontal flip. An example image is depicted in Figure 5.

\begin{figure}[t]
\begin{center}
  \includegraphics[width=0.6\linewidth]{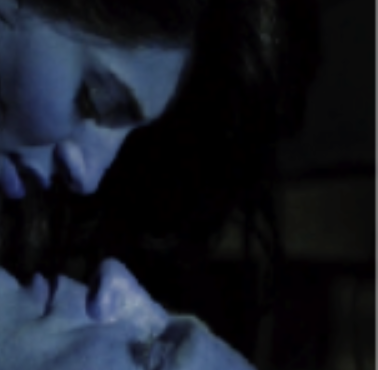}
\end{center}
   \caption{Example of an input image to the image feature extractor}
\label{fig:long}
\label{fig:onecol}
\end{figure}

The extraction logic for ResNet 3D is very similar. We extract 3-channel patches of the same size and then stack up the tensors for up to 16 preceding frames. Missing frames (for the few 15 frames) in a video are filled with zeros. For example the 16x3x224x224 tensor for the first data point in a movie consists of 15 zero tensors of 3x224x224 followed by a tensor containing the actual frame data.

\subsubsection{Audio features}
The audio features are extracted from the last 960ms of the audio wave. The first step is to resample the mono audio wave at 16KHz. We then calculate a short-time Fourier transform, applying a 25ms window at every 10ms interval. The resulting spectogram is integrated into 64 mel-spaced frequency bins (covering 125-7500Hz) and the magnitude of these bins is log-transformed to produce a 96x64 patch \cite{hershey2017cnn}. 

An instance of the generated patch is visualized using a heatmap in Figure 6.

\begin{figure}[t]
\begin{center}
  \includegraphics[width=0.7\linewidth]{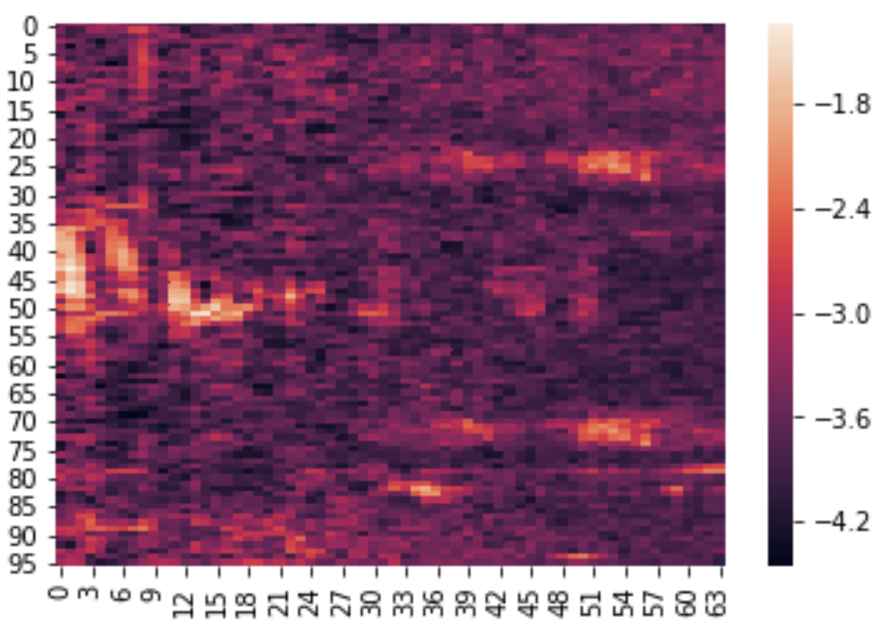}
\end{center}
   \caption{Example of an input into the audio feature extractor}
\label{fig:long}
\label{fig:onecol}
\end{figure}

\section{Experiments}
\subsection{Evaluation metric}
We will use the F1 score to evaluate the quality of the binary classifier. The F1 score is computed as the harmonic mean of precision and recall and strikes a balance between them which makes it harder for the system to cheat by favoring one. By training high quality binary classifiers at the one-second segment level we will be able to have a system that can accurately detect kissing segments. Segmentation is performed by Algorithm 1 as discussed previously.

\subsection{Baseline}
As our baseline we are using a 2D ResNet pre-trained on ImageNet. We froze the parameters in all of the layers except for the last and fine-tuned this model for 15 epochs on our dataset. The validation F1 stays consistent at 0.6 while the training F1 climbs up and approaches 0.8. We also trained the same model from scratch with the same configuration. Validation F1 score as a function of number of epochs is depicted in Figure 9. We will evaluate our proposed classifier against this baseline.

\begin{figure}[t]
\begin{center}
  \includegraphics[width=0.8\linewidth]{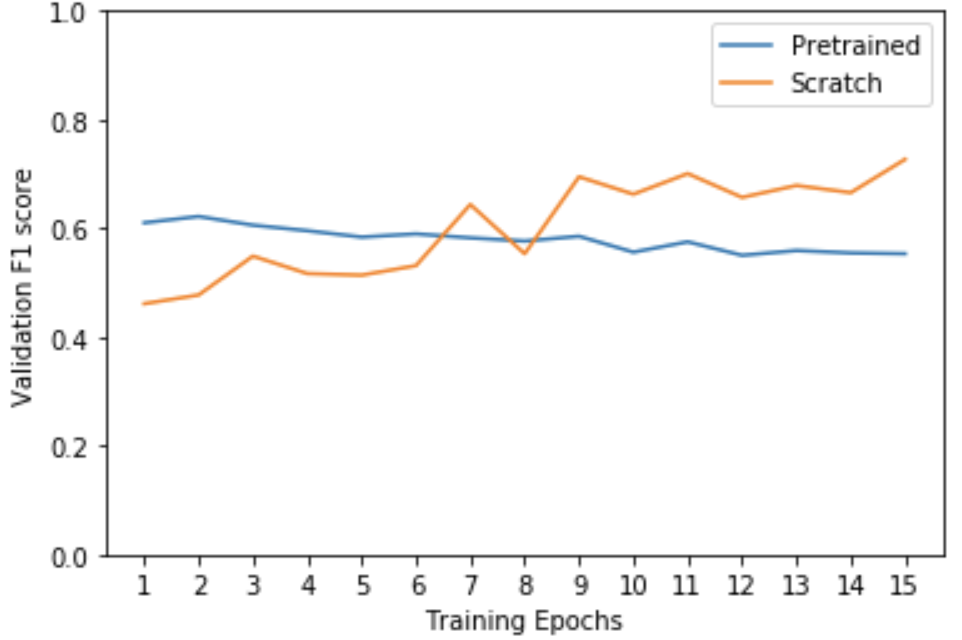}
\end{center}
   \caption{Baseline evaluation}
\label{fig:long}
\label{fig:onecol}
\end{figure}

\subsection{Binary classifier}
We trained our binary classifier over 10 epochs and achieved an evaluation F1 score of 0.95. For this initial experiment we allowed all of the weights in the network to be trained. We used a batch size of 64, ResNet-18 as the image feature extractor, VGGish as the audio feature extractor, and an Adam optimizer with learning rate 0.001 as described earlier.

\subsubsection{Qualitative error analysis}
We used saliency maps to gain some intuition into what the image feature extractor piece of our system has learned about kissing. Saliency maps are computed by taking the gradient of the unnormalized score for the correct class with respect to each pixel in the image which produces is a 3x224x224 tensor. Finally We take the maximum of the absolute value across all channels for each pixel \cite{simonyan2013deep}.

Figure 7 depicts the saliency maps for two successful classifications and one for a non-kissing scene mispredicted as one.

\begin{figure}[t]
\begin{center}
  \includegraphics[width=0.9\linewidth]{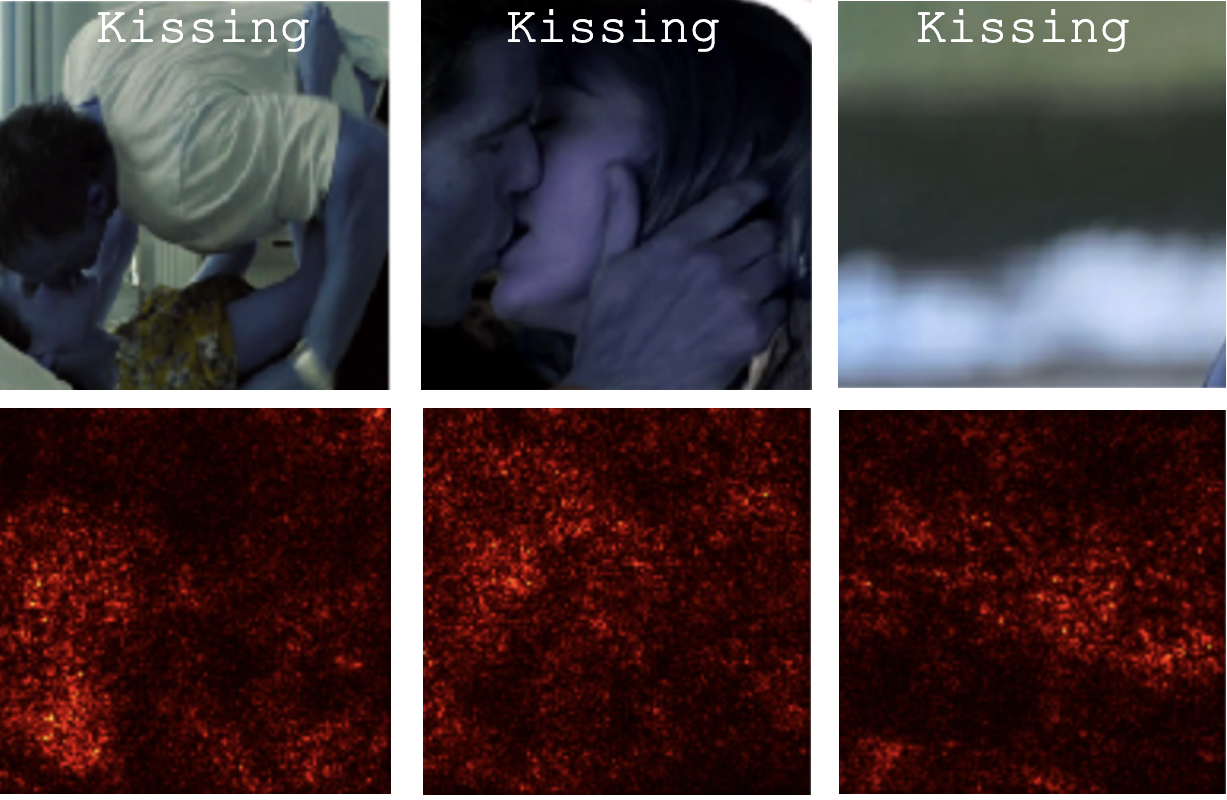}
\end{center}
   \caption{Saliency maps}
\label{fig:long}
\label{fig:onecol}
\end{figure}

The successful predictions seem to indicate that the model has learned about focusing on pixels in the faces. The unsuccessful prediction can be traced to two source of issues arising from the training data. The first source is the random cropping of images. We made no effort in our annotation to only include kissing scenes where the kissing action is in the center of the shot. Some of the kissing scenes in our dataset are wide shots where the scenery dominates the shot and random crops are more likely to be shots of scenery than of actual people.

The second related issue is that some kissing scenes can be very fast-paced and include many cuts where not both of the actors are present briefly. Also in many cases a romantic scene may involve some kissing followed by long stretches of conversation.

These issues highlight two points. First, one needs to construct a more carefully crafted dataset to try to get around some of these issues. Second, capturing contextual information is critical for detecting kissing since still images don't tell the whole story.

\subsection{Ablation study}
In order to understand the contribution of the image and audio feature extraction components we conducted an ablation study.

Table 1 shows the validation F1 score for the three possible combinations repeated with training either all the parameters in the network or just the last fully-connected layer (freezing the rest). The results suggest that training all the parameters with both components present is the best option. We can also conclude that the image feature extraction component has a higher contribution. In other words the system's performance degrades more as we remove the image feature extractor but not so much if the VGGish network, which operates on audio features, is removed.
\begin{table}
\begin{center}
\begin{tabular}{llll}
\hline
Trained                  & VGGish & ResNet-18 & Validation F1 \\ \hline
Last layer               & Yes    & Yes       & 0.92          \\
Last layer               & No     & Yes       & 0.92          \\
Last layer               & Yes    & No        & 0.87          \\
All                      & Yes    & Yes       & \textbf{0.95}          \\
All                      & No     & Yes       & 0.91          \\
All                      & Yes    & No        & 0.87         
\end{tabular}
\end{center}
\caption{Effect of image and feature extraction components}
\end{table}

\subsection{Architecture and hyper-paramter search}
We swept over the values $\{ 0.005, 0.001, 0.01 \}$ for learning rate, architectures \{ DenseNet, ResNet-18 \}, and trained either all parameters or just the final layer. We chose DenseNet  \cite{huang2017densely} as it tends to be more efficient in terms of the number of parameters and computation for the same level of accuracy. Similar to ResNets a DenseNet adds shortcut connections but different from a ResNet it receives the outputs from all previous layers. However, this concatenation of previous layer inputs results in a larger memory footprint for DenseNet.

We tabulate the results for our search in Table 2. The results suggest that ResNet is the superior architecture for this task and training configuration. We also observe that training all the parameters tends to perform better for both architectures and the choice of learning parameter is important and can be further tuned.

\begin{table}
\begin{center}
\begin{tabular}{llll}
\hline
Architecture             & Trained & learning rate & F1 \\ \hline
ResNet-18               & Last layer    & 0.0005       & 0.89          \\
ResNet-18               & Last layer    & 0.001       & 0.92          \\
ResNet-18               & Last layer    & 0.01         & 0.90          \\
ResNet-18                      & All    & 0.0005       & 0.88          \\
ResNet-18                      & All     & 0.001       & \textbf{0.95}          \\
ResNet-18                      & All    & 0.01        & 0.90          \\
DenseNet               & Last layer    & 0.0005       & 0.79          \\
DenseNet               & Last layer    & 0.001       & 0.86          \\
DenseNet               & Last layer    & 0.01         & 0.81          \\
DenseNet                      & All    & 0.0005       & 0.83          \\
DenseNet                      & All     & 0.001       & 0.88          \\
DenseNet                      & All    & 0.01        & 0.88          
\end{tabular}
\end{center}
\caption{Learning rate and architecture search}
\end{table}

\subsection{3D ResNet}
With the best set of hyper-parametesr achieved from the previous step we trained a 3D ResNet-34 (described in an earlier section) for 10 epochs. Training time per epochs was increased by a factor of 10 (i.e. 10 minutes per epoch) and we halved the batch size to 32 for faster training on a p2.xlarge Amazon Web Services (AWS) instance with a single NVIDIA K80 GPU.

The best validation F1 score using this architecture was 0.88, which underperforms ResNet-18. We offer two potential reasons for this finding. First, the ResNet-18 model benefits from being pre-trained on ImageNet while we trained our 3D ResNet from scratch. We attempted pre-trained weights from a model \cite{hara2018can} trained on the Kinetics dataset \cite{kay2017kinetics} to no avail. Because of this reason we probably need to train this model for a substantially larger number of epochs. Second, the temporal depth of the model at 16 frames may not be sufficient to capture the relevant context.

\section{Discussion and future work}
We have designed a system for detecting kissing scenes in a movie. This system relies on a binary classifier that takes audio and image features and produces a binary label. Our experiments suggest that both audio and image features are benefitial for this classification and we observed that the model appears to have learned to pay attention to pixels in the image that are likely an indicator of kissing (e.g. faces).

We also observed that imperfect annotated scenes can confuse the model. The typical transformations applicable to image classification tasks may not be entirely appropriate for this task. For example it may be beneficial to run instance segmentation on the still frames and crop within those regions or to use directly input instance segmentation information to the classifier.

As we alluded to in the previous section we can experiment with training 3D ConvNets for longer and search over hyper-parameters such as temporal depth. These networks are a more promising area to explore as they can, in theory, learn the context and we don't need to spend too much time on perfecting the training set for a detector.

\section{Acknowledgements}
The dataset used in this work and some computing resources were generously provided by Professor Kayvon Fathalian's lab. The 3D ResNet code used in this work is adapted from \href{https://github.com/kenshohara/3D-ResNets-PyTorch}{https://github.com/kenshohara/3D-ResNets-PyTorch}. The VGGish PyTorch architecture and pre-processing is adapted from \href{https://github.com/harritaylor/torchvggish}{https://github.com/harritaylor/torchvggish}. Code for saliency maps is adapted from \href{http://cs231n.github.io/assignments2019}{http://cs231n.github.io}.

{\small
\bibliographystyle{ieee}
\bibliography{egbib}
}

\end{document}